\documentclass[a4paper,fleqn]{cas-dc}
\usepackage{cite}
\usepackage[numbers]{natbib}
\usepackage{cite}
\usepackage{amsmath,amssymb,amsfonts}
\usepackage{graphicx}
\usepackage{textcomp}
\usepackage{xcolor}
\usepackage{indentfirst}

\usepackage{geometry}
\usepackage{algorithmicx}  
\usepackage{algpseudocode}  
\usepackage{amsmath}
\usepackage{subcaption}
\usepackage{amsmath}
\usepackage{enumitem}
\usepackage{caption}
\usepackage{multirow} 
\usepackage{longtable} 
\usepackage{array} 
\usepackage{makecell}
\usepackage{threeparttable} 
\usepackage{booktabs}

\def\tsc#1{\csdef{#1}{\textsc{\lowercase{#1}}\xspace}}
\tsc{WGM}
\tsc{QE}
\tsc{EP}
\tsc{PMS}
\tsc{BEC}
\tsc{DE}
\graphicspath{{figures/}} 
\begin{document}
\let\WriteBookmarks\relax
\def\floatpagepagefraction{1}
\def\textpagefraction{.001}
\shorttitle{Rethinking Spatio-Temporal Transformer for Traffic Prediction:
Multi-level Multi-view Augmented Learning Framework}
\shortauthors{Jiaqi Lin et~al.}
\title [mode = title]{Rethinking Spatio-Temporal Transformer for Traffic Prediction:
Multi-level Multi-view Augmented Learning Framework}

\author[1]{Jiaqi Lin}
\ead{linjiaqi@s.hlju.edu.cn}

\author[1]{Qianqian Ren}  
\cormark[1]
\ead{renqianqian@hlju.edu.cn}
\address[1]{Department of Computer Science and Technology, Heilongjiang University, Harbin, 150080, China}



\cortext[1]{Corresponding author.}


\begin{abstract}
Traffic prediction is a challenging spatio-temporal forecasting problem that involves highly
complex spatio-temporal correlations. This paper proposes a Multi-level Multi-view Augmented
Spatio-temporal Transformer (LVSTformer) for traffic prediction. The model aims to capture
spatial dependencies from three different levels: local geographic, global semantic, and pivotal nodes, along with long- and short-term temporal dependencies. Specifically, we design three spatial augmented views to delve into the spatial information from the perspectives of local, global, and pivotal nodes. By combining three spatial augmented views with three parallel spatial self-attention mechanisms, the model can comprehensively captures spatial dependencies at different levels. We design a gated temporal self-attention mechanism to effectively capture long- and short-term temporal dependencies. Furthermore, a spatio-temporal context broadcasting module is introduced between two spatio-temporal layers to ensure a well-distributed allocation of attention scores, alleviating overfitting and information loss, and enhancing the generalization ability and robustness of the model.
A comprehensive set of experiments is conducted on six well-known traffic benchmarks, the experimental results demonstrate that LVSTformer achieves state-of-the-art performance compared to competing baselines, with the maximum improvement reaching up to 4.32\%.
\end{abstract}


\maketitle

\section{Introduction}
Traffic prediction has become an essential component of Intelligent Transportation Systems (ITS), which encompasses various applications such as traffic management\cite{nellore2016survey}, route planning\cite{bast2016route} and congestion avoidance\cite{jacobson1988congestion}. 
The main challenge lies in efficiently capturing the complex and time-varying spatio-temporal dependencies of traffic data.
Recurrent Neural Networks (RNNs)\cite{rumelhart1986learning,cho2014learning} and their variants, such as LSTM\cite{zhang2016dnn} and GRU\cite{fu2016using}, are used to capture temporal dependencies of traffic data. Nonetheless, these methods fail to model spatial correlations. To address this limitation, recent research has
combined Convolutional Neural Networks (CNNs)\cite{lecun1998gradient,krizhevsky2012imagenet,szegedy2015going} and RNNs to capture spatio-temporal dependencies of grid-based traffic data, with models like ST-ResNet[37] and STDN[34] proposed for this purpose. However, CNNs have inherent limitations in handling common non-Euclidean data representations.
Recently, Spatio-Temporal Graph Neural Networks (STGNNs) have been developed for traffic prediction. These models combine GNNs with either RNNs or Temporal Convolutional Networks (TCNs) to capture the spatio-temporal correlations of traffic data. Noteworthy STGNN models incldue DCRNN\cite{li2017diffusion}, STGCN\cite{yu2017spatio}, AGCRN\cite{bai2020adaptive}, TGCN\cite{zhao2019t}, GTS\cite{shang2021discrete} and GMSDR\cite{liu2022msdr}.

Building upon the success of attention mechanisms and Transformers in various deep learning fields, including natural language processing (NLP)\cite{vaswani2017attention}, and image 
\begin{figure}
\centerline{\includegraphics[width=0.45\textwidth]{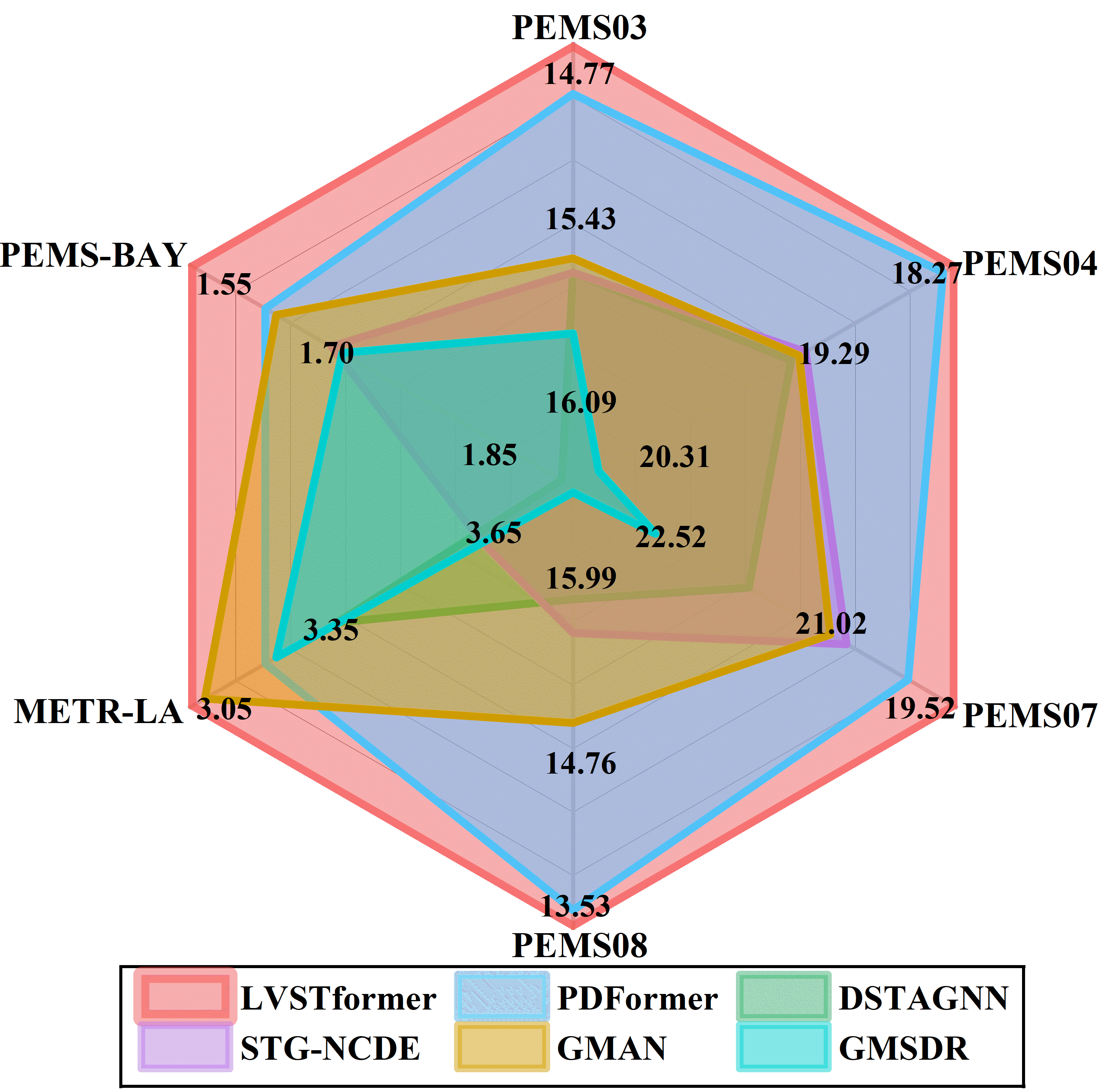}}
\caption{Performance comparisons with respect to MAE on six traffic datasets. Our LVSTformer achieves the best performance.}
\label{figs/intro}
\end{figure}
classification\cite{dosovitskiy2020image}, several studies have integrated attention mechanisms into the field of time series forecasting, resulting in significant performance improvements in capturing complex dependencies and enhancing prediction accuracy. For example, models like Informer\cite{zhou2021informer}, Pyraformer\cite{liu2021pyraformer} and PatchTST\cite{nie2022time} have achieved promising performance in time series forecasting tasks. However, these models were not specifically designed for traffic prediction and do not adequately consider complex spatial dependencies, which lead to suboptimal performance in traffic prediction. Models such as ASTGCN\cite{guo2019attention}, GMAN\cite{zheng2020gman}, 
DSTAGNN\cite{lan2022dstagnn} and PDFormer\cite{jiang2023pdformer} have successfully introduced attention mechanisms into  traffic prediction. These models leverage transformers, attention mechanisms, and hybrid architectures to make predictions, while considering both temporal and spatial dependencies. Consequently, they have achieved surprising results in capturing complex spatio-temporal patterns.
Nevertheless, there are still challenges to overcome:

\begin{figure*}
\centerline{\includegraphics[width=1.0\textwidth]{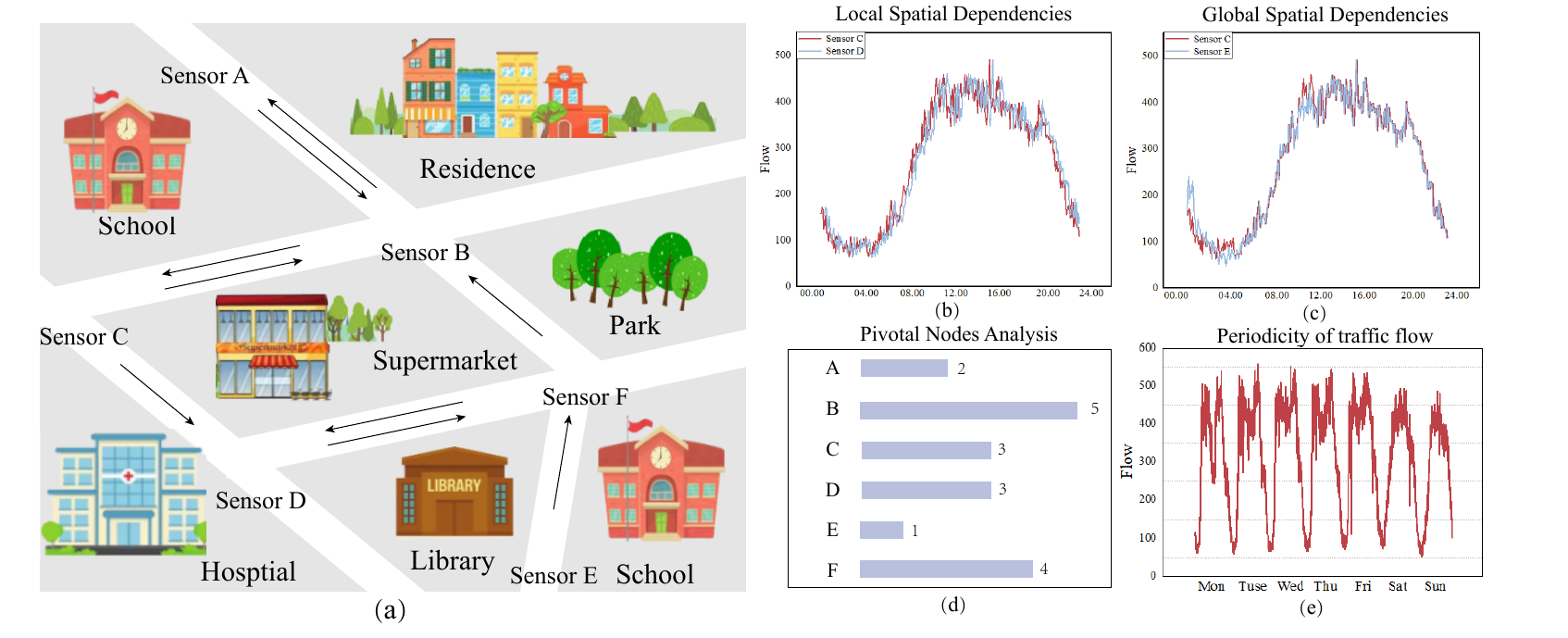}}
\caption{Figure (a) illustrates the region division and sensors deployment, while figures (b) and  (c) respectively demonstrate the local and global spatial dependencies in traffic data. Figure  (d) counts the input and output flows of each node, and figure (e) showcases the periodicity of traffic flow.}
\label{figs/example}
\end{figure*}

 \textbf{Multi-level spatial correlations.} Figure \ref{figs/example}(b) demonstrates that geographically adjacent nodes are influenced by same regional factors tend to demonstrate similar traffic patterns. Additionally, regions that are functionally similar but distant may also exhibit similar traffic patterns. For example, in Figure \ref{figs/example}(c), Sensor C and Sensor F are far apart, yet both are located near a school, indicating functional similarity, and thus they demonstrate similar traffic patterns.
Furthermore, within traffic networks, certain nodes exhibit more intricate spatial dependencies due to their extensive connections with multiple other nodes, exemplified by Sensor B in Figure \ref{figs/example}(a). These nodes are termed pivotal nodes. Pivotal nodes demonstrate superior abilities in aggregating and distributing traffic flow compared to other nodes.
Specifically modeling the features of pivotal nodes and allocating corresponding computational weights based on their abilities to aggregate and distribute traffic flow contribute to enhancing predictions of overall traffic patterns and network behavior. Most methods treat spatial correlations as a whole, ignoring multi-level spatial correlations present.

\textbf{Long- and short-term temporal dependencies and periodicity.} Although STGNNs utilizing RNN or TCN to model short-term temporal correlations have been proven effective, it is difficult for them to obtain satisfactory results for long-term prediction. With the increasing traffic pressure in cities, there is an urgent need for methods that can effectively model long- and short-term temporal dependencies to meet the requirements of efficient traffic management and planning.
Furthermore, we have found significant periodic trends in traffic patterns, as shown in Figure \ref{figs/example}(e). Each day, during peak morning and evening hours, traffic flow significantly increases, while it decreases during the night. Capturing and embedding these periodic trends into the model can better model temporal dependencies.
However, existing methods often overlook the importance of periodicity, resulting in errors when modeling temporal dependencies.

\textbf{Attention imbalance.} In attention-based models, different input features are assigned different scores to help the model identify and process information. However, there is a risk of excessively high or low attention scores being assigned to certain features. 
This imbalance can result in the model overly relying on specific features while disregarding other valuable information, thereby reducing generalization capability and performance.

To tackle the previously mentioned issues, we propose LVSTformer, a Multi-level $\&$ Multi-view Augmented Spatio-Temporal Transformer framework for traffic prediction, which contains three key components: \textbf{spatio-temporal data embedding layer}, \textbf{multi-view generation} and \textbf{multi-level spatio-temporal transformer}.
The spatio-temporal data embedding layer serves as the initial processing stage, where we aggregate raw traffic data, temporal features, and spatial features. This integration effectively models the spatio-temporal features of traffic data, providing comprehensive representation for subsequent processing.
The multi-level spatio-temporal transformer layers form as the core of the model, enabling effective capture of complex spatio-temporal dependencies. Specifically, in the temporal dimension, we utilize gated temporal self-attention to capture temporal correlations, which enhances the extraction of both local and global temporal features, thereby augmenting the model's perception of local environmental information and long-term traffic patterns.
In the spatial dimension, LVSTformer holds advantages over existing transformer-based approaches, as we propose three parallel spatial self-attention mechanisms and integrate three spatial enhanced views into these mechanisms to capture spatial dependencies from multiple levels (local geographic, global semantic, and pivotal nodes). To address the issue of attention imbalance in attention-based models, we introduce the Spatio-Temporal Context Broadcasting (STCB) module. This module manually inserts uniform attention between two layers of the model, promoting a relatively balanced distribution of attention. The proposed LVSTformer demonstrates the best performance across six real-world traffic datasets, as illustrated in Figure \ref{figs/intro}.
The paper's contributions can be outlined as follows:

\textbf{General Aspect.} 
We emphasize the importance of capturing both long- and shot-term temporal dependencies, along with multi-level spatial dependencies, in traffic prediction tasks. Hence, 
we introduce a Multi-level $\&$ Multi-view Augmented Spatio-Temporal transformer framework (LVSTformer) to effectively  capture intricate spatio-temporal dependencies. 
In addition, we alleviate the issue of attention imbalance through the STCB module.

\textbf{Methodologies.} 
    First, we embed traffic raw data, temporal periodic information, and spatial information into the model through a spatio-temporal data embedding layer. Next, we introduce a multi-level multi-view augmented transformer network, which integrates multi-level spatial self-attention mechanisms for local geographic, global semantic, and pivotal nodes, along with a gated temporal self-attention mechanism. Then, we separately integrate the information from three spatial enhanced views into the spatial self-attention mechanism, facilitating enhanced capture of spatial features from different perspectives by the model. Finally, the spatio-temporal context broadcasting module allows for proper adjustment and optimization of attention, ensuring a reasonable distribution of attention scores and thereby enhancing the model's generalization capability and robustness.

\textbf{Experiments Evaluation.} We conduct extensive experiments on six traffic datasets to evaluate the performance of LVSTformer and compare it with state-of-the-art baseline methods. The results show that LVSTformer is highly competitive.

\section{Related Work}
\subsection{Traffic Prediction }
Traditional models such as HA\cite{campbell2008predicting}, 
ARIMA\cite{zhang2003time}, and VAR\cite{stock2001vector} have been employed for traffic prediction. However, these methods face difficulties in grasping the complex spatio-temporal relationships inherent to traffic information. Subsequently, machine learning methods such as KNN\cite{guo2003knn} and SVR\cite{wu2004travel} can capture non-linear dependencies, but they require manual selection and
expert intervention, making them unsuitable for processing large-scale traffic data. As computational power improves and deep learning continues to evolve, STGNNs, Attention Mechanism and Transformers have emerged as the dominant technologies.

\subsection{Spatio-Temporal Graph Neural Networks(STGNNs)}
STGNNs combine GNNs with RNNs and TCNs, demonstrating powerful capabilities in capturing spatio-temporal dependencies and significantly outperforming traditional
statistical models. For instance, DCRNN\cite{li2017diffusion} utilizes diffusion convolutional networks and GRU to capture spatio-temporal dependencies, while STGCN\cite{yu2017spatio} utilizes GCN and GRU to model spatio-temporal correlations. GWNET\cite{wu2019graph} uses stacked multilayer gated temporal convolutions with GCN units, notably designing an adaptive matrix to account for the 
interactions between neighboring nodes.
 AGCRN\cite{bai2020adaptive} employs two enhanced GCNs to capture spatiotemporal correlations respectively between specific nodes and different sequences. GMSDR\cite{liu2022msdr} introduces a novel variant of recurrent neural networks: Multi-Step Dependency Relations (MSDR). Based on this, it seamlessly integrates GCN with MSDR to effectively model spatio-temporal dependencies.
Nevertheless, due to inherent limitations in RNNs and TCNs, while they excel at capturing short-term temporal correlations, they often struggle to model long-range temporal correlations in long sequences.

\subsection{Attention Mechanism and Transformer}
Due to Transformer's notable performance in capturing spatio-temporal correlations and its ability to effectively alleviate challenges such as gradient explosion and vanishing, it has become one of the primary model architectures in the field of spatio-temporal prediction. The attention mechanism is the central component of the Transformer architecture, which models dependencies between a query and a set of values by adaptively assigning weights, determined by the queries and associated keys 
within the set. In this setup, each symbol in the input sequence interacts with
all other symbols, creating an effective global receptive field. Recent studies 
have shown that attention mechanisms and transformers excel in the field of traffic forecasting, such as ASTGCN\cite{guo2019attention} effectively combines GCN and attention mechanisms to model spatio-temporal features. DSTAGNN\cite{lan2022dstagnn} and PDFormer\cite{jiang2023pdformer} not only utilize attention mechanisms to capture spatiotemporal correlations, but also respectively consider historical traffic flow data for dynamic inter-node attributes and traffic delay effects.

\subsection{Miscellaneous}
In addition to STGNNs and STtransformers, some new traffic prediction methods have been proposed. For example, STFGNN\cite{li2021spatial} uses dynamic time warping (DTW) to construct time graphs based on time series similarity, while STSGCN\cite{song2020spatial} connects individual spatial graphs of adjacent time steps to construct local spatio-temporal graphs. Furthermore, methods based on differential equations, such as STGODE\cite{fang2021spatial} and STG-NCDE\cite{choi2022graph}, which combine GCNs with differential equations, are achieving promising results.
\begin{table}[htbp]
\caption{Summary of existing models.}\label{result1}
\centering
\scalebox{0.7}{
\begin{tabular}{c|c|c|ccc|ccc}
\toprule

\multicolumn{1}{c}{} &\multicolumn{1}{c}{} &\multicolumn{1}{c}{}  & \multicolumn{3}{c}{Spatial} & \multicolumn{3}{c}{Temporal}  \\ 

\multicolumn{1}{c}{Methods} &\multicolumn{1}{c}{Conference} &\multicolumn{1}{c}{Year}   & \multicolumn{1}{c}{CNN} & \multicolumn{1}{c}{GCN} & \multicolumn{1}{c}{Attn.}  & \multicolumn{1}{c}{RNN} & \multicolumn{1}{c}{TCN} & \multicolumn{1}{c}{Attn.} \\

\hline
ST-ResNet   &AAAI   &2017		&\checkmark&	    &		& &   &	 \\
DCRNN  	    &ICLR   &2018		&&\checkmark	    &		&\checkmark &   &	 \\
STGCN       &IJCAI  &2018		&&\checkmark	    &		& &\checkmark   & \\
STDN    &AAAI   &2019		&\checkmark&	    &		&\checkmark &   &\checkmark	 \\
GWNET       &IJCAI	&2019		&&\checkmark	    &		& &\checkmark   & \\
TGCN        &IEEE TITS	&2019	&	&\checkmark	    	&	&\checkmark &   & \\
ASTGCN 	    &AAAI	&2019		&&\checkmark	    &\checkmark		& &\checkmark   &\checkmark \\
STSGCN  	&AAAI	&2020		&&\checkmark	    &		& &   & \\
AGCRN		&NIPS	&2020		&&\checkmark	    &		&\checkmark &   & \\
GMAN		&AAAI	&2020		&&    &\checkmark			& &   &\checkmark	 \\
GTS		&ICLR	&2021		&&\checkmark	    &		&\checkmark &   & \\
GMSDR		&KDD	&2022	&	&\checkmark	    &		&\checkmark &   & \\
DSTAGNN	    &ICML	&2022	&	&\checkmark	    &\checkmark		& &\checkmark   &\checkmark \\
PDFormer    &AAAI   &2023	&	&	    &\checkmark		& &   &\checkmark	 \\
\bottomrule
\end{tabular}}
\end{table}

\section{Problem Formulation}
 We denote the road network as a graph  $\mathbf{G} = (\mathbf{V},\mathbf{E},\mathbf{A})$, where $\mathbf{V}=\{v_i\}_{i=1,2,\cdots,N}$ represents the set of nodes, which represent sensors within the road network. $\mathbf{E}=\{e_{ij}\}$ denotes the set of edges, denoting the connection between pairs of nodes. The adjacency matrix $\mathbf{A}\in \{0,1\}^{N\times N}$ is used to represent whether the corresponding road segments are connected,
 for $\forall v_i,v_j\in \mathbf{V}$, if $e_{ij} \in \mathbf{E}$, then $\mathbf{A}_{ij}$
 is set to 1, indicating a connection between the corresponding road segments.  Otherwise $A_{ij}$ is set to 0. The feature matrix $\mathbf{X}\in \mathbb{R}^{N\times T}$  describes the traffic features, where $T$ and  $N$ represents the number of time steps and nodes respectively. $\mathbf{x}_t \in\mathbb{R}^{N}$ represents the features of all nodes at time $t$.

\textbf{Problem Statement} (Traffic Foretasting): In traffic forecasting, given the historical traffic feature observations $\mathbf{X}=(\mathbf{x}_{t-T+1}, $
$\mathbf{x}_{t-T+2},...,\mathbf{x}_{t-1}, \mathbf{x}_t) \in \mathbb{R}^{N\times T}$ with $T$ time steps and $N$ nodes, our objective is to forecast the future $T'$ time steps $\mathbf{Y}=(\mathbf{x}_{t+1},\mathbf{x}_{t+2},...,\mathbf{x}_{t+T'-1}, $
${X}_{t+T'}) \in \mathbb{R}^{N\times T'}$. 

\begin{figure*}
\centerline{\includegraphics[width=1.00\textwidth]{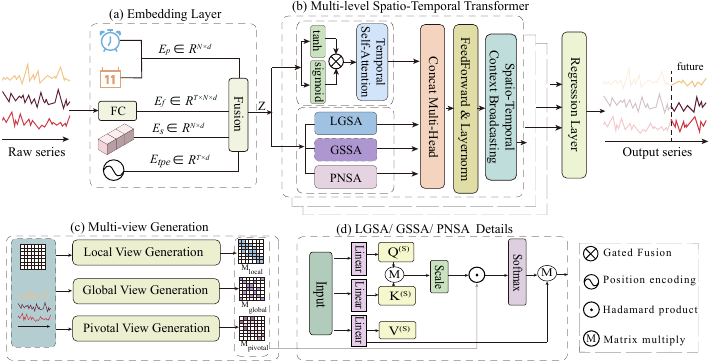}}
\caption{
The architecture of the LVSTformer:
(a) Embedding Layer aggregates raw traffic data, temporal periodic features, and spatial features to effectively model the spatio-temporal features of traffic data.
(b) Multi-level Spatio-Temporal Transformer captures temporal dependencies through the gated self-attention, and spatial dependencies through spatial self-attention, which consists of three modules, local geographic self-attention(LGSA), global
semantic self-attention(GSSA), and pivotal nodes self-attention(PNSA). 
(c) Multi-view Generation constructs local view, global view, and pivotal view, which are integrated with spatial self-attention.
(d) The details of LGSA, GSSA and PNSA, they share the same architecture.}
\label{figs/framework}
\end{figure*}

\section{METHODOLOGY}
Figure.\ref{figs/framework} illustrates the overall framework of LVSTformer, which is composed of three major components: the spatio-temporal data embedding layer, the multi-view generation module and the multi-level spatio-temporal Transformer.

\subsection{Spatio-Temporal Embedding Layer}
To enhance the capabilities of the model backbone, we have introduced an embedding layer between the original input and the transformer encoder. The embedding layer operates on the input data from three main perspectives: 
(1) converting raw traffic data into high-dimensional representations.
(2) embedding periodic knowledge and position encoding for transformer-based models.
(3) employing graph laplacian to embed spatial information.

\subsubsection{Raw Data Embedding}
In order to preserve the original information in the road network, we employ fully connected layers to generate raw data embedding denoted as $\mathbf{E}_f \in \mathbb{R}^{T \times N \times d}$, where $d$ represents the dimension of the hidden layer. $\mathbf{E}_f$ is computed as follows:
\begin{equation}
\mathbf{E}_f = \text{FC}(\mathbf{x}_{t-T+1:t})
\end{equation}
where the fully connected layer $\text{FC}(\cdot)$ processes the input sequence $\mathbf{x}_{t-T+1:t}$, transforming it into a feature embedding that captures essential information while retaining the inherent characteristics of the original data. This feature embedding serves as a rich representation for subsequent components of the model to effectively leverage and process the preserved raw information.

\subsubsection{Temporal Periodic Embedding}
To incorporate the periodicity of traffic flow in the temporal dimension, we introduce two transformation functions: $f_w(t)$ and $f_d(t)$. These functions are utilized to recognize and transform the time $t$ into the corresponding day of the week (ranging from 1 to 7) and time of the day (ranging from 1 to 1440) as follows:
\begin{equation}
\mathbf{E}_d = f_d(t),\quad \mathbf{E}_w = f_w(t)
\end{equation}

To create temporal periodic embeddings, we concatenate $\mathbf{E}_d$ and $\mathbf{E}_w$ across all time steps. The temporal periodic embeddings $\mathbf{E}_p \in \mathbb{R}^{T \times d}$ is thus obtained.

\subsubsection{Temporal Position Encoding}
After obtaining the temporal periodic embedding $\mathbf{E}_p$, the next step is to perform temporal position encoding. Due to the significant temporal relative positional relationships exhibited by traffic data, the position encoding aims to capture and incorporate the relative positional information of the sequence into the original Transformer framework.
The position encoding scheme assigns unique positional embeddings to each element in the sequence based on their relative positions.
Specifically, the positional embedding procedure is formulated as follows: 
\begin{equation}
    \begin{aligned}
\text{$PE_{pos}(i)$}=\begin{cases}sin(p/10000^{i/d}),&\text{if $i$ is even }\\ cos(p/10000^{(i-1)/d}),&\text{if $i$ is odd}\end{cases}
\end{aligned}
\end{equation}
where $d$ represents dimension of embedding, $p$ is position index. The temporal position encoding is denoted as $\mathbf E_{tpe} \in \mathbb R^{T \times d}$.
By using this formulation, the positional embedding captures the relative positional relationships within the temporal dimension. The sine and cosine functions introduce oscillating patterns that encode the position information, allowing the model to differentiate between different positions in the sequence.

\subsubsection{Spatial Graph Laplacian Embedding}
In the spatial dimension, the graph Laplacian eigenvectors are considered for capturing the graph structure of the road network. The Laplacian eigenvectors are computed based on the input graph, represented as:

\begin{equation}
\mathbf L=\mathbf I-\mathbf D^{-1/2}\mathbf{AD}^{-1/2}=\mathbf U^\mathbf {T}\Lambda U
\end{equation}
where $\mathbf{A}$ is the adjacency matrix, $D$ is the diagonal degree matrix of $\mathbf{A}$ with $D_s(i,j)=\mathbf\sum_{j=1}^{n}\mathbf{A}_{(ij)}$, $\mathbf I$ is the identity matrix. By performing eigenvalue decomposition, we acquire $\mathbf \Lambda$ and $\mathbf U$, where $\mathbf \Lambda$ is the eigenvalue matrix satisfying $0=\lambda_0<\lambda_1\leq...\leq\lambda_{N-1}$ and $\mathbf U=(u_0,u_1,...,u_{N-1})$ is the eigenvector matrix.

To create the spatial positional embedding, we utilize the $k$ smallest non-trivial eigenvectors from $\mathbf{U}$ for each node in the graph. These eigenvectors capture the spatial information associated with the node's position within the road network graph.
Given a node in the graph, we concatenate the $k$ eigenvectors corresponding to that node to form a spatial positional embedding of dimension $k \times 1$. This process will be repeated for each node in the graph, while applying a learnable linear projection to the positional embedding, in order to generate the final spatial graph Laplacian embedding $\mathbf{E}_s$ of dimensions $N \times d$, where $N$ represents the number of nodes in the graph.


\subsubsection{Data Embedding Output}
Finally, by concatenating the four kinds of embeddings mentioned above, the final output is represented as $\mathbf{Z}$:
\begin{equation}
    \mathbf Z= \mathbf E_f||\mathbf E_p||\mathbf E_s||\mathbf E_{tpe}
\end{equation}
Here, $||$ represents the concatenation operation. By concatenating these embeddings, the resulting $\mathbf{Z}$ has a combined representation of the spatio-temporal features of the traffic data. 

\subsection{Multi-view Generation}\label{mlvg}
In order to  comprehensively explore the spatial dependency from different perspectives in traffic prediction tasks, we generated three spatial augmentation views corresponding to three types of spatial dependencies: (1) Local spatial dependency, which captures the similarity among nearby nodes. (2) Global spatial dependency, which captures the similarity among nodes located in similar functional regions. (3) Pivotal spatial dependency, which 
investigates the influence and dependencies associated with specific nodes that play a critical role in the transportation network.
This approach aims to provide the model with a broader spatial perspective, enabling it to capture and understand complex spatial dependencies in traffic prediction tasks effectively.
In the remaining part of this section, we will elaborate on how to generate these three spatial augmentation views: the local view, the global view, and the key node view.

\textbf{Local View:}
To construct the local view, we utilize the shortest path algorithm to calculate the distances between nodes in the network. Let $D(i,j)$ represent the distance from node $i$ to node $j$. The calculation of the shortest path distances can be described as follows:
\begin{equation}
D(i,j) = \min(D(i,j), D(i,k) + D(k,j))
\end{equation}
In this equation, we iterate over all possible intermediate nodes $k$ and update the distance matrix $D$ by considering all pairs of nodes $i$ and $j$. The algorithm aims to find the minimum distance between any two nodes by either directly traversing an edge between $i$ and $j$ or by going through an intermediate node $k$.

Once we have obtained the shortest paths between all pairs of nodes, we can set a threshold based on practical considerations to identify which nodes are considered neighbors and should be included in the local view. 
If the shortest distance from nodes $i$ and $j$ is less than the threshold, the corresponding entry in $\mathbf{M}_{local}(i,j)$ is set to 1, otherwise 0:
\begin{equation}
\mathbf M_{local}(i, j) = 
\begin{cases} 
1 & \text{if } D(i, j) < \text{threshold} \\
0 & \text{otherwise}
\end{cases}
\end{equation}
In this way, we can eliminate the influence of nodes that are far apart, thereby focusing attention on neighboring nodes.

\textbf{Global View:}
We employ $DTW$\cite{sakoe1978dynamic} to calculate the similarity of historical traffic data(flow or speed) between nodes in the network. First, we calculate the daily average traffic data for each node. This step is crucial as it helps reduce the noise and variability present in the raw traffic data. Next, we use the $DTW$ algorithm to compute the similarity of data between node $i$ and all other nodes.
 \begin{equation}
Similarity(i,j)= DTW(\text{AVG}(X_{d,i}), \text{AVG}(X_{d,j}))
\end{equation}
where $j \in \{1, \ldots, N\}$, $i\neq j$. $\text{AVG}(X_{d,i})$ and $\text{AVG}(X_{d,j})$ represent the daily average of node $i$ and $j$. 
we select the $K_g$ nodes that exhibit the highest similarity to node $i$ as its global similarity neighbors. We define the set of global similarity neighbors for node $i$, denoted as $C_i$, as follows:
 \begin{equation}
C_i = \{j|Similarity(i,j) \in TopK_g(Similarity)\}
\end{equation}
Following the same procedure for all nodes in the network, we can obtain the set of global similarity neighbors for each node. Finally, we construct the global masking vector $\textbf M_{global}$. Specifically, if node $i$ and node $j$ are mutual global neighbors, then $\textbf M_{global}(i,j)$ = 1. Otherwise, the entry is set to 0. By traversing all possible node pairs, we construct the binary global view matrix $\textbf M_{global}$.
\begin{equation}
\mathbf M_{global}(i, j) = 
\begin{cases} 
1 & \text{if }j \in C_i \wedge i \in C_j   \\
0 & \text{otherwise}
\end{cases}
\end{equation}
By considering the similarity of nodes' functions and traffic patterns, the model is capable of capturing long-range global spatio-temporal dependencies within the network.


\textbf{Pivotal View: }
In a road network, certain nodes exhibit stronger abilities in aggregating and distributing traffic. We refer to these nodes as pivotal nodes. We quantify the aggregation and distribution capabilities of all nodes in the road network to identify pivotal nodes.
Specifically, we begin by extracting the OD matrix, denoted as $\mathbf{M_{(o,d)}}$, from the original dataset, where $o$ represents the origin region and $d$ represents the destination region. Each entry $\mathbf{M_{(o,d)}}$ records the connectivity from a specific origin region to a specific destination region. Nodes appearing more frequently as origin regions indicate stronger capabilities in distributing traffic, while nodes appearing more frequently as destination regions indicate stronger capabilities in aggregating traffic.
Subsequently, we propose a scoring function to quantify the aggregation and distribution capabilities of nodes. Taking node 
$i$ as an example:
\begin{equation}
    \begin{aligned}
&Score(i)= \sum_{j=1}^{N}(\mathbf{M}_{(i,j)} + \mathbf{M}_{(j,i)})\\
\end{aligned}
\end{equation}
After obtaining the scores for each node, we define the set of pivotal nodes $P$ by selecting the $K_p$ nodes with the highest scores. The set $P$ is defined as follows:
 \begin{equation}
P = \{i|Score(i) \in TopK_p(Score)\}
\end{equation}
where $K_p$ is a hyperparameter that determines the number of key nodes to be selected. 
Once the set of pivotal nodes is obtained, we can construct the pivotal weight matrix $\textbf{M}_{{pivotal}}$ to assign weights based on the score values. 
\begin{equation}
\textbf M_{{pivotal}}(i, j) = 
\begin{cases} 
Score(i)+Score(j) & \text{if }i \in P \vee j \in P   \\
0 & \text{otherwise}
\end{cases}
\end{equation}
If both $i$ and $j$ are pivotal nodes, then the value of $\textbf{M}_{{pivotal}}(i,j)$
is the sum of their scores. For nodes where both are non-pivotal or where one of them does not belong to the pivotal node set, their weights are set to 0, indicating that they do not possess the same level of importance in traffic dynamics.

\subsection{Multi-level Spatio-Temporal Transformer}
The architecture of the Multi-level Spatio-Temporal Transformer(MLST) module is illustrated in Figure \ref{figs/framework}(b), inspired by the encoder of the transformer model.
We first introduce
the detailed design of our multi-view spatial attention module in Section 4.3.1, and the temporal attention module in Section 4.3.2. These modules jointly capture spatial and temporal dependencies within the data. Next, we introduce multiple attention fusion in Section 4.3.3. Finally, the spatio-temporal context broadcasting technique is proposed in Section 4.3.4.

\subsubsection{Multi-view Spatial Self-Attention Module}\label{sam}
To exploit the spatial features at different levels, we further propose a Multi-view Spatial Self-Attention(MVSA) module, which consists of three parallel attention mechanisms: Local Geographic Self-Attention(LGSA), Global Semantic Self-Attention(GSSA), and Pivotal Nodes Self-Attention(PNSA). These mechanisms are designed to capture the spatial dependencies of local, global, and pivotal nodes, respectively. Since LGSA, GSSA, and PNSA share the same architecture, we will explain the details using LGSA as an example. As shown in Figure \ref{figs/framework}(d),
given an input $Z$ at time $t$, a 1 × 1 convolutional operation is applied to obtain the query ${{\mathbf{Q}}}^{(S)}_{t,h}$, the key ${{\mathbf{K}}}^{(S)}_{t,h}$, and the value ${{\mathbf{V}}}^{(S)}_{t,h}$: 

\begin{equation}
    \begin{aligned}
&\mathbf{Q}^{(S)}_{t,h}={W}_{t,h}^{(Q)}\mathbf{Z}, \quad &\mathbf{K}^{(S)}_{t,h}={W}_{t,h}^{(K)}\mathbf{Z}, \qquad
 &\mathbf{V}^{(S)}_{t,h}={W}_{t,h}^{(V)}\mathbf{Z}
\end{aligned}
\end{equation}
where ${W}_{t,h}^{(Q)}$, ${W}_{t,h}^{(K)}$, ${W}_{t,h}^{(V)}\in\mathbb{R}^{d\times d'}$ are learnable parameters, $d'$ is the dimension of the query, key, and value matrices and $h \in \{1, \ldots, h_l\}$, $h_l$ is the number of heads in LGSA. The multi-head attention is given by:

\begin{equation}
\begin{aligned}
&head_{t,h}(\mathbf{Q}^{(S)}_{t,h},\mathbf{K}^{(S)}_{t,h},\mathbf{V}^{(S)}_{t,h},\mathbf{M}_{local}) \\
&= \left. \mathrm{softmax}\left(\frac{\mathbf{Q}^{(S)}_{t,h}\left(\mathbf{K}^{(S)}_{t,h}\right)^\top}{\sqrt{d^{^{\prime}}}}\odot \mathbf{M}_{local}\right)\mathbf{V}^{(S)}_{t,h} \right.
\end{aligned}
\end{equation}
\begin{equation}
H_{local}=Concate(head_{t,1},head_{t,2},\cdots, head_{t,h_l})W^l
\end{equation}
Here $W^l$ is a learnable parameter for the output.
Similarly, by integrating the attention mechanism with $\mathbf{M}_{global}$ and $\mathbf{M}_{pivotal}$ respectively, we can obtain the outputs of GSSA and PNSA, denoted as $H_{global}$ and $H_{pivotal}$. Through this approach, the MVSA module can effectively capture spatial information at multiple levels without incurring additional computational costs or requiring complex operations. Specifically, LGSA focuses on capturing spatial dependencies within a local neighborhood. GSSA considers capturing spatial dependencies between nodes with similar functions across the entire road network, 
which are usually long-distance. PNSA emphasizes the importance of pivotal nodes, as these nodes have a stronger ability to aggregate and distribute traffic flow and often exhibit more complex spatial dependencies.

\subsubsection{Gated Temporal Self-Attention Module}\label{tam}
Transformers are known for their effectiveness in handling long sequence modeling, but it fails to effectively learn the inherent trend features present in temporal variations.
To address this limitation, we propose a gated temporal self-attention to capture both short- and long-term temporal features in traffic data. The module incorporates a gated network composed of filters and gates. This gating mechanism enables the model to dynamically regulate the flow of information and selectively emphasize the relevant trend features while filtering out noise or irrelevant information. The gated temporal self-attention is given by:
\begin{equation}
 \begin{aligned}
&filter = \mathrm{tanh}(\mathrm{Conv}(\mathbf{Z})),\\
&gate = \mathrm{sigmoid}(\mathrm{Conv}(\mathbf{Z})),\\
&\bar {\mathbf Z}= filter \odot gate
\end{aligned}
\end{equation}
The tanh and sigmoid functions serve as activation functions, and 
$\bar {\mathbf Z}$ is input for the subsequent operations. Next, we utilize
 the temporal self-attention to obtain the dynamic features of local and global among all time slices, as well as various temporal trends. 
Specifically, for node
$n$, we first obtain the query matrix $\mathbf{Q}^{(T)}_{n,s}$, the key matrix $\mathbf{K}^{(T)}_{n,s}$, and the value matrix $\mathbf{V}^{(T)}_{n,s}$. 
\begin{equation}
    \begin{aligned}
&\mathbf{Q}^{(T)}_{n,s}=\mathbf{W}_{n,s}^{(Q)}\bar {\mathbf Z};\quad
&\mathbf{K}^{(T)}_{n,s}=\mathbf{W}_{n,s}^{(K)}\bar {\mathbf Z};\qquad
&\mathbf{V}^{(T)}_{n,s}=\mathbf{W}_{n,s}^{(V)}\bar {\mathbf Z}
\end{aligned}
\end{equation}
where $\mathbf{W}_{n,s}^{(Q)}$, $\mathbf{W}_{n,s}^{(K)}$, $\mathbf{W}_{n,s}^{(V)}\in\mathbb{R}^{d\times d'}$  are learnable parameters. $s \in \{1, \ldots, h_t\}$, $h_t$ is the number of heads in GTSA.
 
Subsequently, we employ self-attention operations to reveal the dependencies among all time slices for node $n$.
Formally, the temporal attention is described as follows:
\begin{equation}
head_{n,s}(\mathbf{Q}^{(T)}_{n,s},\mathbf{K}^{(T)}_{n,s},\mathbf{V}^{(T)}_{n,s})=\mathrm{softmax}\left(\frac{\mathbf{Q}^{(T)}_{n,s}\left(\mathbf{K}^{(T)}_{n,s}\right)^\top}{\sqrt{d^{^{\prime}}}}\right)\mathbf{V}^{(T)}_{n,s}\\
\end{equation}

\begin{equation}
H_{t}=Concate(head_{n,1},head_{n,2},\cdots, head_{n,h_t})W^t
\end{equation}


\subsubsection{Multi-head Attention Fusion}\label{maf}
To enhance computational efficiency, the outputs of four types of attention mechanisms, ${H}_{local}$, ${H}_{global}$, ${H}_{pivotal}$, and ${H}_{t}$, are integrated into a multi-head attention block, which is represented as follows:
\begin{equation}
\begin{split}
H&=Concat({H}_{local},{H}_{global},{H}_{pivotal},{H}_{t})\\
{\hat {\mathbf{Z}}}&=Linear(H)
\end{split}
\end{equation}

We follow the original Transformer model by applying feed-forward network and layer normalization to the output of the multi-head attention to acquire the output of the layer, with the result still denoted as 
 ${\hat {\mathbf{Z}}}\in\mathbb{R}^{T\times N\times d}$.

\subsubsection{Spatio-Temporal Context Broadcasting}\label{stcb}
\begin{figure*}
\centerline{\includegraphics[width=0.60\textwidth]{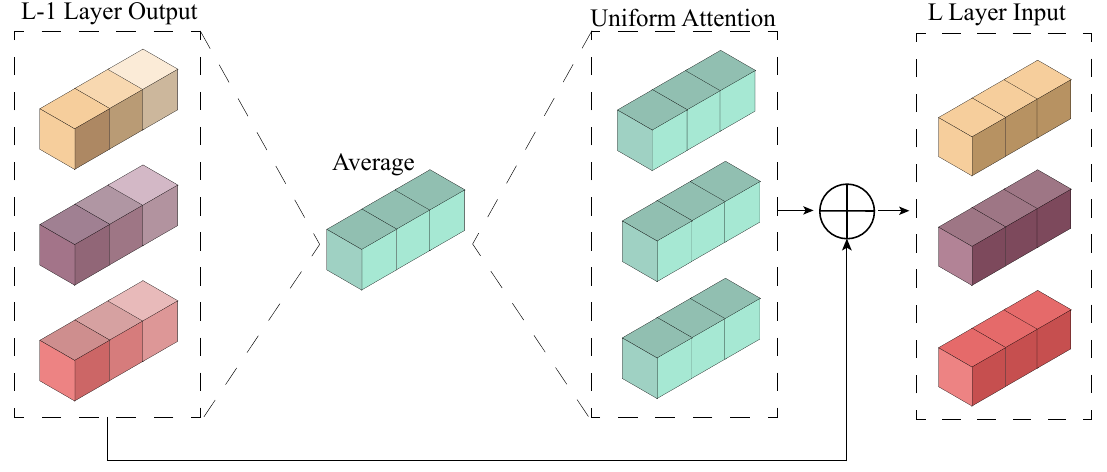}}
\caption{The structure of STCB.}
\label{figs/stcb}
\end{figure*}

In order to prevent excessively high or low attention scores, which may lead the model to focus excessively on certain features or neglect other valuable information, 
we need a method to ensure that the model can consider various features more comprehensively. Inspired by \cite{hyeon2023scratching}, we introduce a complementary operation called the Spatio-Temporal Context Broadcasting (STCB), which inserts uniform attention manually to ensure a reasonable distribution of attention scores. Specifically, it is represented as:

\begin{equation}
\begin{split}
    STCB(\hat z_{i}) &= \frac{\hat z_{i}+\frac{1}{N}\sum_{j=1}^{N}\hat z_{j}}{2}
\end{split}
\end{equation}
where $\hat z_i$ is the $i^{th}$ token in ${{\hat{\mathbf{Z}}}}\in\mathbb{R}^{T\times N\times d}$. Figure \ref{figs/stcb} depicts the design of STCB, positioned at the end of the MLP, following the sequence of FC Layer1 -> GeLU -> FC Layer2 -> STCB. 
To determine the optimal placement of the STCB module within the MLP, we conducted experiments by placing it in different positions. Specifically, we tried placing the STCB module in front of FC Layer1, behind FC Layer1, and behind FC Layer2. The experimental results indicate that the optimal performance is attained when the STCB module is placed at the end of FC Layer2.

Consider such a traffic scenario where a significant event occurs in a specific area of the city, resulting in significant congestion. Without the STCB module, the model may excessively focus on observations within that specific area, accurately capturing the traffic congestion there. However, it may inadvertently overlook observations from other relatively important areas of the city, resulting in suboptimal predictions for those regions. 
By introducing the STCB module, this issue can be alleviated to some extent, allowing the model to distribute attention more evenly to other relatively important areas, thereby broadening the model's perspective. As far as we know, our team is the first to tackle this particular challenge in traffic prediction.

\subsection{Regression Layer}
After the previous processing steps, the output $\hat{\mathbf{Z}}$ is further processed using two $1\times1$ convolutional layers to transform it to the desired output dimension. This is denoted as follows:
\begin{equation}
\mathbf{Y}=\mathrm{Conv}_{2}(\mathrm{Conv}_{1}(\hat{\mathbf{Z}}))
\end{equation}
where $\mathrm{Conv}_{1}$ and $\mathrm{Conv}_{2}$ represent two $1*1$ convolutional layers. The result $\mathbf{Y}\in \mathbb R^{T'\times N \times C}$ represents the prediction results for $T'$ time steps.

\begin{table*}[h]
\caption{{Dataset description and statistics.}}
\centering
\scalebox{1.1}{
\begin{tabular}{cccccccc}
\toprule
Datasets  &\#Nodes   &\#Edges   &\#Time Steps   &\#Time Interval     &\#Time Period  &\#Type  \\
\midrule
PeMS03     &358      &574      &26208         &5 minutes         &2018/09-2018/11 &Flow \\
PeMS04     &307      &340      &16992         &5 minutes         &2018/01-2018/02 &Flow \\
PeMS07     &883      &866      &28224         &5 minutes         &2017/05-2017/08 &Flow \\
PeMS08     &170      &295      &17856         &5 minutes         &2016/07-2016/08 &Flow \\
PeMS-BAY   &325      &2369     &52116         &5 minutes         &2017/01-2017/06 &Speed \\
METR-LA    &207      &1515     &34272         &5 minutes         &2012/03-2012/06 &Speed\\
\bottomrule
\end{tabular}}\label{dataset}
\end{table*}
\section{EXPERIMENT}
 We seek to answer the following inquiries in our research:
\begin{itemize}
 \item \textbf{RQ1}: How does LVSTformer perform compared with SOTA methods?
 \item \textbf{RQ2}: How does each component influence LVSTformer's performance? 
  \item \textbf{RQ3}: How does LVSTformer perform in addressing long-term prediction?
  \item \textbf{RQ4}: How do different hyperparameters setting affect the
 prediction performance of LVSTformer?
   \item \textbf{RQ5}: How does the computation cost of LVSTformer compare to that of other models?
 \end{itemize}

\subsection{Experimental Settings}

\subsubsection{Datasets Details}
We perform extensive experiments on six real-world traffic datasets. 
The specific details of the six datasets are shown in Table \ref{dataset}. Here are brief introductions to these six datasets:
\begin{itemize}
  \item  PeMS03: It comprises traffic flow measurements from District 3 in California, USA and encompasses data from 358 sensors, spanning from Sep. 2018 to Nov. 2018.

  \item PeMS04: It comprises traffic flow measurements from District 4 in California, USA, collected from Jan. 1 to Feb. 28, 2018, totaling data from 307 sensors.

\item PeMS07: It comprises traffic flow measurements from District 7 in California, USA, collected from May 1, 2017, to Aug 6, 2017, from a total of 883 sensors.

\item  PeMS08: It comprises traffic flow measurements from District 8 in California, USA. It consists of data from 170 sensors, collected from Jul 1, 2016, to Aug 31, 2016.

\item METR-LA: It collects traffic speed data from various regions of Los Angeles, California, USA, spanning from Mar 1, 2012, to Jun 27, 2012, and comprises readings from a total of 207 sensors.

  \item PeMS-BAY: It collects traffic speed measurements from various regions of the San Francisco Bay Area, California, USA, covering from Jan 1, 2017, to Jun 30, 2017, and comprises readings from a total of 325 sensors.

\end{itemize}
The traffic flow dataset and the traffic speed dataset are split into training, validation, and test sets with ratios of 6:2:2 and 7:1:2, respectively.

\begin{table*}[htbp]
\caption{Comparison results of different approaches on PeMS03, PeMS04 and PeMS07 Datasets.}\label{result1}
\centering
\scalebox{1.15}{
\begin{tabular}{ccccccccccc}
\toprule

\multicolumn{1}{c}{} & \multicolumn{3}{c}{PeMS03} & \multicolumn{3}{c}{PeMS04} & \multicolumn{3}{c}{PeMS-07}  \\ 
\cline{2-10} 
\multicolumn{1}{c}{Methods}  & \multicolumn{1}{c}{MAE} & \multicolumn{1}{c}{RMSE} & \multicolumn{1}{c}{MAPE} & \multicolumn{1}{c}{MAE} & \multicolumn{1}{c}{RMSE} & \multicolumn{1}{c}{MAPE} & \multicolumn{1}{c}{MAE} & \multicolumn{1}{c}{RMSE} & \multicolumn{1}{c}{MAPE}  \\

\hline
HA      &31.58&52.39&33.78\%&38.03 &59.24 &27.88\%  &45.12 &65.64  &24.51\%  \\
ARIMA     &35.41&47.59&33.78\%     	&33.73	&48.80	&24.18\%	&38.17	&59.27	&19.46\%	\\
VAR      &23.65&38.26&24.51\%      	&24.54	&38.61	&17.24\% 	&50.22	&75.63	&32.22\% 	 \\
SVR    &20.73&34.97&20.63\%                &27.23      &41.82     &18.95\%        &32.49      &44.54      &19.20\%   \\
\cline{1-10}
FC-LSTM   &21.33&35.11&23.33\%                  &26.77      &40.65     &18.23\%        &29.98      &45.94      &13.20\%                          \\
DCRNN   	&17.99&30.31&18.34\%&21.22	&33.44	&14.17\%	&25.22	&38.61	&11.82\%	\\
STGCN	&17.55&30.42&17.37\% 	&21.16	&34.89	&13.83\%	&25.33	&39.34	&11.21\%	 \\
ASTGCN &17.34&29.56&17.21\%	&22.93	&35.22	&16.56\%	&24.01	&37.87	&10.73\%	 \\
STSGCN  &17.48&29.21&16.78\%	&21.19	&33.65	&13.90\%	&24.26	&39.03	&10.21\%	\\
AGCRN	&15.98&28.25&15.23\%	&19.83	&32.26	&12.97\%	&22.37	&36.55	&9.12\%	 \\
STFGNN	&16.77&28.34&16.30\%	&20.48	&32.51	&16.77\%	&23.46	&36.60	&9.21\%	 \\
GMAN	&15.52&26.53&15.19\%	&19.25	&30.85	&13.00\%	&20.68	&33.56	&9.31\%	 \\
STGODE	&16.50&27.84&16.69\%	&20.84	&32.82	&13.77\%	&22.59	&37.54	&10.14\%	\\
STG-NCDE &15.57&27.09&15.06\%		&19.21	&31.09	&12.76\%	&20.53	&33.84	&8.80\%	 \\
DSTAGNN	 &15.57&27.21&\textbf{14.68\%}&19.30	&31.46	&12.70\%	&21.42	&34.51	&9.01\%	 \\
GMSDR    &15.78&\underline{26.82}&15.33\%&20.49	&32.13	&14.15\%	&22.27	&34.94	&9.86\%	\\
PDFormer &\underline{14.95}&27.22&15.39\%   &\underline{18.36}	&\underline{30.03}	&\underline{12.10\%}	&\underline{19.97}	&\underline{32.95}	&\underline{8.55\%}	 \\
\cline{1-10}
\textbf{LVSTformer(Ours)}
&\textbf{14.77}&\textbf{26.71}&\underline{14.96\%}
&\textbf{18.27}&\textbf{30.01}&\textbf{12.00\%}
&\textbf{19.52}&\textbf{32.56}&\textbf{8.42\%}
\\
\bottomrule
\end{tabular}}
\end{table*}
\subsubsection{Baselines}
To assess the proposed model, we selected 17 baseline models for comparison. These baseline models include a range of traditional statistical models, spatio-temporal graph neural network models, attention-based models, and some other models. Here is a brief description of these baseline models:
\begin{itemize}
    \item  HA\cite{campbell2008predicting}: It utilizes historical averages of traffic flow to forecast future traffic conditions.

    \item  ARIMA\cite{zhang2003time}: It combines auto-regressive, moving average, and differencing components to model and forecast data..

    \item  VAR\cite{stock2001vector}: It can infer the causal relationships and dynamic influences among multiple variables.

    \item  SVR\cite{wu2004travel}: It uses support vector regression model to predict upcoming traffic conditions.
    
    \item FC-LSTM\cite{sutskever2014sequence}: It effectively models temporal correlations by leveraging the advantages of both fully connected layers and LSTM.
    
    \item  DCRNN\cite{li2017diffusion}: It combines diffusion convolutional network and RNN to model and predict data.

    \item  STGCN\cite{yu2017spatio}: It combines GCN and ST convolution to effectively model the spatio-temporal correlations.

    \item  ASTGCN\cite{guo2019attention}: It introduces attention mechanism for adaptive adjustment of node and edge weights, thereby more effectively capturing crucial information.
    \item  STSGCN\cite{song2020spatial}: It utilizes synchronous convolution, enabling the model to more effectively extract spatio-temporal features.

    \item  AGCRN\cite{bai2020adaptive}: It can adaptively learn patterns of specific nodes and infer the interdependencies of data across different time periods.
    \item  STFGNN\cite{song2020spatial}: It generates adaptive spatio-temporal fusion maps with DTW to efficiently learn hidden spatio-temporal dependencies.

    \item  GMAN\cite{zheng2020gman}: It captures spatiotemporal features through an encoder, and the decoder is used to generate prediction results.

    \item  STGODE\cite{fang2021spatial}: It uses ODE to model temporal dynamics and GNN to capture spatial dependencies.

    \item  STG-NCDE\cite{choi2022graph}: It processes continuous sequences through NCDE to learn finer temporal dynamics and uses GNN to handle the spatial structure in the data.

    \item  DSTAGNN\cite{lan2022dstagnn}: It combines GNN with attention to flexibly capture the crucial features of spatio-temporal data.

    \item  GMSDR\cite{liu2022msdr}: This model uses the fusion mechanism of temporal and spatial attention to comprehensively model long-distance and multi-scale spatio-temporal patterns.


    \item  PDFormer\cite{jiang2023pdformer}: It combines propagation delay awareness and dynamic long-range transformer to handle dynamic and complex traffic data.
\end{itemize}
\begin{table*}[htbp]
\caption{Comparison results of different approaches on PeMS08, METR-LA and PeMS-BAY Datasets.}\label{result2}
\centering
\scalebox{1.15}{
\begin{tabular}{ccccccccccc}
\toprule

\multicolumn{1}{c}{} & \multicolumn{3}{c}{PeMS08} & \multicolumn{3}{c}{METR-LA} & \multicolumn{3}{c}{PeMS-BAY}  \\ 
\cline{2-10} 
\multicolumn{1}{c}{Methods}  & \multicolumn{1}{c}{MAE} & \multicolumn{1}{c}{RMSE} & \multicolumn{1}{c}{MAPE} & \multicolumn{1}{c}{MAE} & \multicolumn{1}{c}{RMSE} & \multicolumn{1}{c}{MAPE} & \multicolumn{1}{c}{MAE} & \multicolumn{1}{c}{RMSE} & \multicolumn{1}{c}{MAPE}  \\

\hline
HA    &34.89 &59.24 &27.88\%  &5.75 &11.34 &13.00\%  &2.88 &5.59  &6.80\%   \\
ARIMA &31.09	&44.32	&22.73\% 	&5.15	&10.45	&12.70\%	&2.33	&4.76 &5.40\%	\\
VAR  &19.19	&29.81	&13.1\%  	&5.28	&9.06	&12.50\% 	&2.24	&3.96 &4.83\%	\\
SVR &22.00      &33.85      &14.23\%   &5.05    &10.87     &12.10\%    &2.48      &5.18      &5.50\% \\
\cline{1-10}
FC-LSTM   &23.09     &35.17    &14.99\%   &3.86   &7.40   &11.70\%    &2.21   &4.57   &5.24\%                          \\
DCRNN  	 &16.82	&26.36	&10.92\%    &3.15	&6.56	&\underline{8.67\%}	    &1.75	&3.92	&3.93\%	 \\
STGCN	 &17.50	&27.09	&11.29\% 	&3.69	&7.43	&9.67\%	    &1.87	&4.58	&4.30\%	 \\
ASTGCN 	&18.25	&28.06	&11.64\%     &3.57	&7.19	&10.32\%	&1.86	&4.07	&4.27\%	 \\
STSGCN  &17.13	&26.80	&10.96\% 	&3.32	&6.66	&9.06\%	    &1.86	&4.03	&4.25\%	 \\
AGCRN	&15.95	&25.22	&10.09\% 	&3.19	&6.41	&8.84\%	    &1.62	&3.61	&\underline{3.66\%}	 \\
STFGNN	&16.94	&26.25	&10.60\% 	&3.18	&6.40	&8.81\%	    &1.66	&3.74	&3.77\%	 \\
GMAN	&14.87	&24.06	&9.77\% 	&\underline{3.08}	&6.41	&\textbf{8.31\%}	    &1.63	&3.77	&3.69\%	 \\
STGODE	&16.81	&25.97	&10.62\% 	&4.73	&7.60	&11.71\%	&1.77	&3.89	&4.02\%	 \\
STG-NCDE 	&15.45	&24.81	&9.92\% 	&3.58	&6.84	&9.91\%	    &1.68	&3.66	&3.80\%	 \\
DSTAGNN	  &15.67	&24.77	&9.94\%   &3.32	&6.68	&9.31\%	    &1.89	&4.11	&4.26\%	\\
GMSDR     &16.36	&25.58	&10.28\%  &3.21	&6.41	&8.76\%	    &1.69	&3.80	&3.74\%\\
PDFormer  &\underline{13.66}	&\underline{23.55}	&\underline{9.06\%}   &3.19	&6.46	&9.29\%	    &\underline{1.62}	&\underline{3.52}	&3.67\%\\

\cline{1-10}
\textbf{ LVSTformer(Ours)}
&\textbf{13.53}&\textbf{23.28}&\textbf{9.05\%}
&\textbf{3.05}&\underline{6.35}&8.81\%
&\textbf{1.55}&\textbf{3.46}&\textbf{3.56\%}
\\
\bottomrule
\end{tabular}}
\end{table*}
\subsubsection{Evaluation Metrics}
We choose three commonly used evaluation metrics in the field of traffic prediction to assess model performance: Mean Absolute Error (MAE), Root Mean Square Error (RMSE), and Mean Absolute Percentage Error (MAPE). The definitions for these metrics are as follows:
\begin{equation}
    MAE = \frac{1}{\Omega}\sum _{i=1} ^{\Omega}  {|\mathbf y^i - \mathbf x^i| }
\end{equation}

\begin{equation}
    RMSE = \sqrt{\frac{1}{\Omega }\sum _{i=1} ^{\Omega}{|\mathbf y^i -\mathbf x^i|^2 }}
\end{equation}

\begin{equation}
    MAPE = \frac{1}{\Omega } \sum _{i=1} ^{\Omega} {|\frac{\mathbf y^i-\mathbf x^i}{\mathbf x^i}|}
\end{equation}
where $\mathbf y$ and $\mathbf x$ are the predicted and actual observed values, respectively.

\subsubsection{Implementation Details}
We conduct experiments on a
server equipped with a GeForce RTX 3090 GPU. In our experiments, we
optimize the model training process using the Adam optimizer. The hidden dimensions are configured to 64, the initial learning rate is set to 0.001, the batch size is set to 16, and the model is trained for a maximum of 200 epochs. For the purpose of prediction, 
we use a historical data window spanning one hour (12 steps) to forecast the data for the following hour.

\subsection{Main Results(RQ1)}
\subsubsection{Comparison Results}
Table \ref{result1} and Table \ref{result2} display prediction results of LVSTformer and baseline methods on six datasets. In the tables, the optimal performance is indicated by values in bold, while suboptimal performance is underlined.
Our LVSTformer model consistently achieves superior performance in most cases. Notably, we observe improvements of up to 1.20\%, 2.25\%, and 4.32\% in MAE for the PeMS03, PeMS07, and PeMS-BAY datasets, respectively. Additionally, Figure \ref{figs/longterm} presents the comparative results between LVSTformer and the top-performing baseline methods on PeMS-Bay and PeMS08 datasets.
 It is evident from the figure that our model exhibits exceptional performance in multi-step prediction. LVSTformer nearly achieves state-of-the-art levels in forecasting for 15, 30, 60 minutes.
Based on these results, we draw the following conclusions:
\begin{figure*}
\centerline{\includegraphics[width=0.8\textwidth]{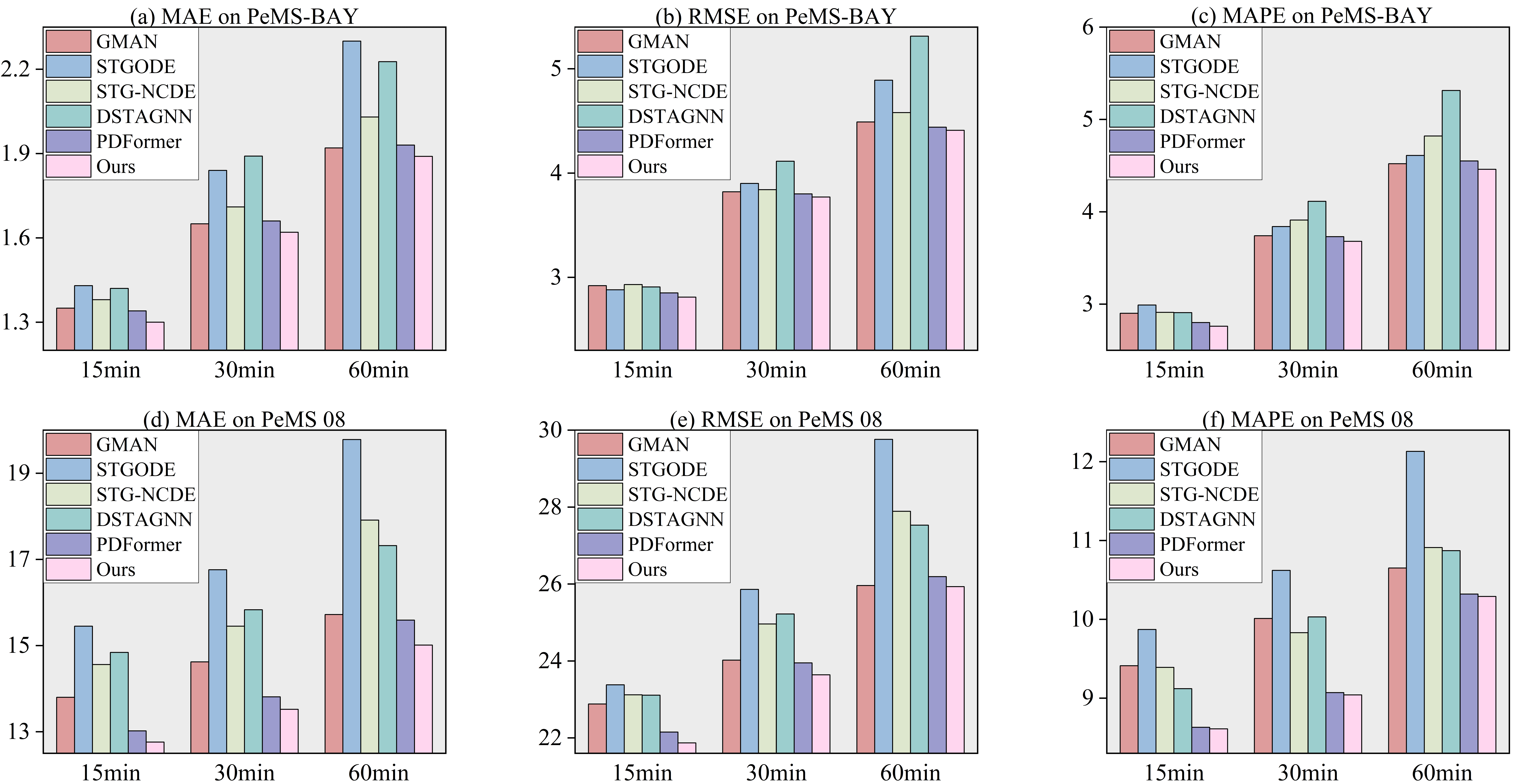}}
\caption{
Comparison results of different methods for multi-step prediction on the PeMS-BAY and PeMS08 datasets.}
\label{figs/longterm}
\end{figure*}
 \begin{figure*}
\centerline{\includegraphics[width=0.80\textwidth]{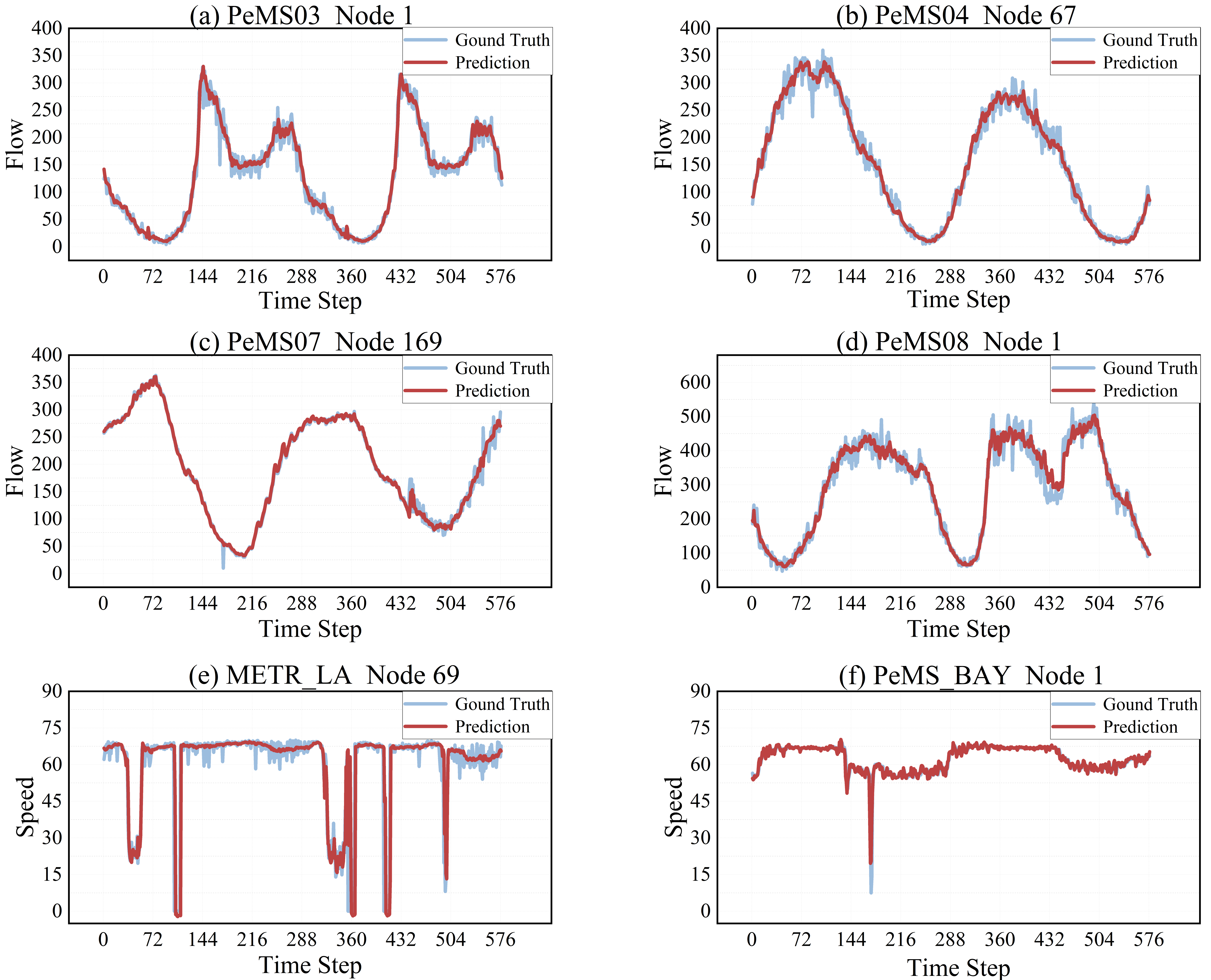}}
\caption{Forecasting visualization of LVSTformer on six datasets.}
\label{可视化}
\end{figure*}
\begin{itemize}
 \item Traditional statistical methods have inherent limitations when dealing with nonlinear and non-stationary data. Consequently, their prediction accuracy tends to be significantly lower compared to deep learning methods.
    
    \item Spatio-temporal GCN based methods, such as DCRNN, AGCRN and GMSDR, can model spatio-temporal correlations to attain better performance. However, when capturing temporal dependencies, these methods are prone to accumulating errors during multi-step prediction. This means that any errors occurring in previous time steps can potentially affect subsequent predictions, thereby resulting in a performance decline. GMSDR has developed a new variant of RNN, which addresses this issue under certain conditions. However, due to the lack of further consideration for spatial correlations, its performance is not the best.
      
    \item Attention-based methods, including GMAN, DSTAGNN and PDFormer, utilize attention mechanisms for modeling spatio-temporal correlations.  GMAN only considers simple spatio-temporal dependencies, DSTAGNN constructs a graph to mine historical data to extract the dynamic inter-node attributes.
    However, these methods do not explore spatial dependencies from multiple scales.
    PDFormer not only makes a simple distinction in spatial attention but also considers the time delay issue in information propagation in real road networks. Hence, it outperforms other models in terms of prediction accuracy.
    \begin{figure*}
\centerline{\includegraphics[width=0.8\textwidth]{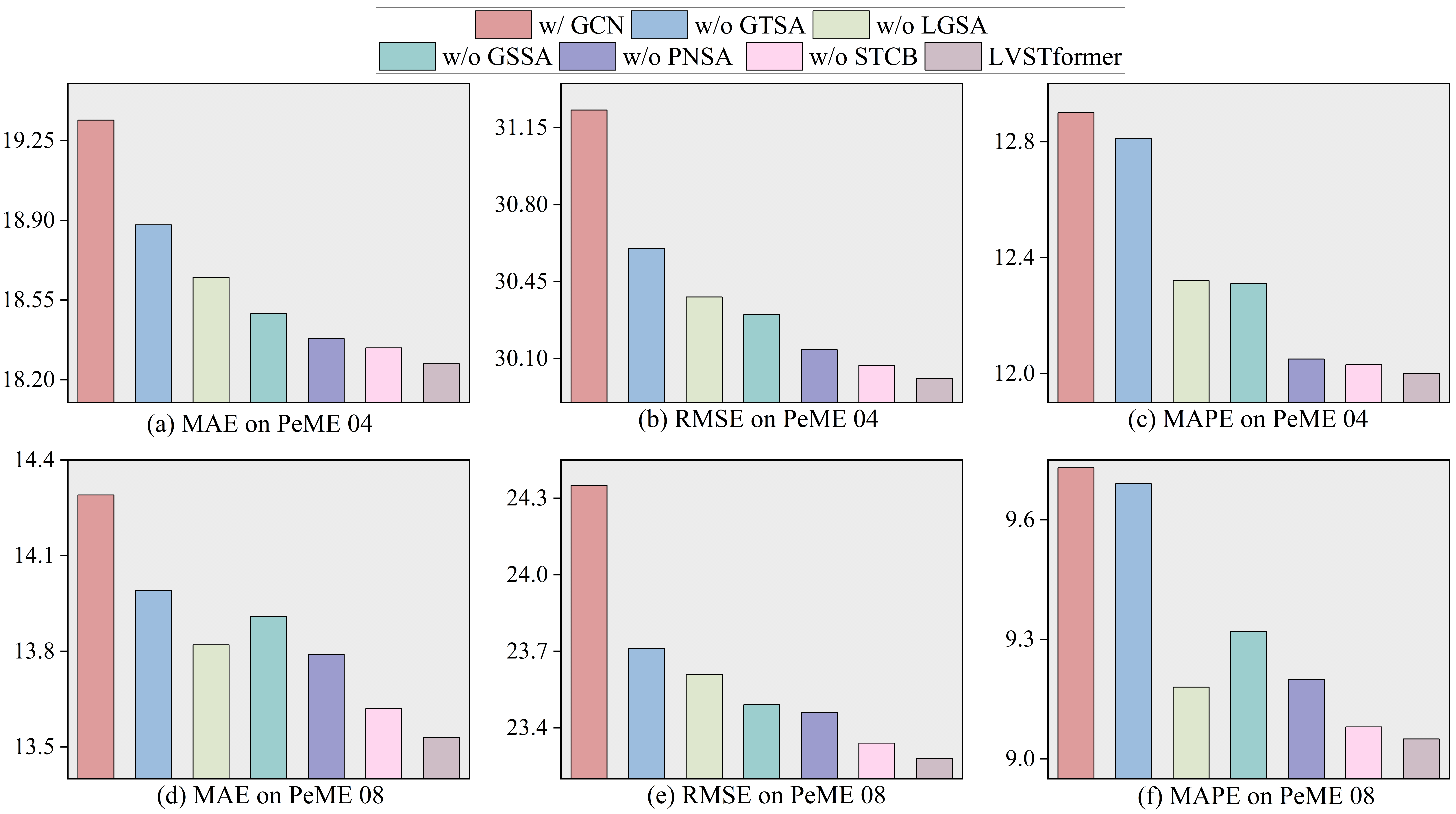}}
\caption{Ablation results on PeMS04 and PeMS08 Datasets.}
\label{figs/curve}
\end{figure*}
        
    \item STGODE and STG-NCDE, utilize differential equations to capture temporal correlations. Compared to RNN, differential equations provide a global view of the entire sequence. They outperform RNN by capturing long-range temporal correlations, but still fall short compared to attention mechanisms.
\end{itemize}
The main reasons for LVSTformer achieving optimal performance are as follows:
\begin{itemize}[itemsep=0 pt, topsep=0 pt]
   \item 
In the data embedding layer, we aggregate raw traffic data, temporal periodic features, and spatial features from the road network, thereby better integrating the spatio-temporal features.
   \item In the spatial dimension, LVSTformer not only captures local and global spatial patterns but also focuses on the spatial dependencies of pivotal nodes within the road network.
In the temporal dimension, LVSTformer enhances the model's ability to capture temporal dependencies by combining a gating network with temporal self-attention.
By capturing multi-level spatial dependencies and long short-term temporal dependencies, the model can comprehensively and accurately analyze and predict spatial data.

   \item  LVSTformer further enhances performance through the STCB mechanism, which reduces the risk of neglecting potentially valuable information obscured by attention.
 \end{itemize}

\subsubsection{Forecasting Results and Visualization}
To illustrate the model’s performance more clearly, one node from each of six datasets is randomly selected for visualization. Figure \ref{可视化} shows the visualization results. By comparing the predicted values from LVSTformer with the ground truth, we observe that prediction curves closely align with actual values. It is evident that LVSTformer promptly responds to dynamic changes in traffic features and accurately captures traffic fluctuations.
Additionally, it observes that LVSTformer exhibits excellent adaptability to short-term drastic changes in traffic data. As shown in Figure \ref{可视化}(e), on METR-LA dataset at $t\in[100,120]$, LVSTformer continues to achieve outstanding prediction accuracy.
Furthermore, our model also shows excellent performance over changing long time intervals, e.g., $t\in[140,280]$ on the PEMS03 dataset.

\subsection{Ablation Study(RQ2)}
In this section, we evaluate the effectiveness of the various components of the model, the definitions of various variants are as follows: 
\begin{table*}
\caption{Long-term prediction performance of LVSTformer on PeMS04 Dataset}
\label{longterm}
\scalebox{0.85}{
\begin{tabular}{lcccccc}
\toprule
\multirow{2}{*}{Model}&\multirow{2}{*}{Metrics} & \multicolumn{4}{c}{Time Interval} & \multirow{2}{*}{Average} \\ 
\cmidrule{3-6}
& & 30 min & 60 min & 90 min & 120 min \\
\midrule
       & MAE & 22.08±0.28 & 25.51±0.69 & 29.32±1.17 & 34.04±1.42 & 26.01±0.75 \\
ASTGCN & RMSE & 34.47±0.42 & 39.35±1.10 & 44.95±1.87 & 54.60±2.20 & 40.64±1.28 \\
       & MAPE (\%) & 14.70±0.10 & 16.84±0.19 & 19.28±0.28 & 22.49±0.31 & 17.22±0.19 \\
\midrule
       & MAE & 21.66±0.36 & 24.04±0.41 & 26.70±0.52 & 29.07±0.64 & 24.35±0.47 \\
STSGCN & RMSE & 34.56±0.75 & 37.98±0.72 & 41.91±0.75 & 45.45±0.90 & 38.46±0.79 \\
       & MAPE (\%) & 14.44±0.13 & 15.76±0.11 & 17.50±0.18 & 18.92±0.15 & 16.13±0.20 \\
\midrule
     & MAE & 20.50±0.01 & 21.02±0.04 & 21.55±0.08 & 22.29±0.05 & 21.08±0.05 \\
GMAN & RMSE & 33.21±0.42 & 34.18±0.48 & 35.09±0.56 & 36.13±0.54 & 34.24±0.49 \\
     & MAPE (\%) & 15.06±0.52 & 15.37±0.57 & 15.78±0.66 & 16.54±0.76 & 15.48±0.60 \\
\midrule
        & MAE & 19.36±0.04 & 20.69±0.08 & 21.69±0.03 & 22.91±0.15 & 20.60±0.02 \\
DSTAGNN & RMSE & 31.36±0.17 & 33.65±0.27 & 35.29±0.22 & 36.81±0.04 & 33.47±0.15 \\
        & MAPE (\%) & 12.88±0.02 & 13.54±0.03 & 14.22±0.01 & 15.04±0.05 & 13.58±0.02 \\
\midrule
       & MAE &\textbf{18.37±0.10} &\textbf{19.44±0.15} &\textbf{20.53±0.12} &\textbf{21.69±0.11} &\textbf{19.52±0.13} \\
 LVSTformer & RMSE &\textbf{30.13±0.07} &\textbf{31.56±0.09} &\textbf{33.02±0.03}&\textbf{34.28±0.08} &\textbf{31.62±0.06} \\
       & MAPE (\%) &\textbf{12.12±0.14} &\textbf{12.89±0.10} &\textbf{13.52±0.13} &\textbf{14.56±0.12} &\textbf{13.09±0.11} \\
\bottomrule
\end{tabular}}
\end{table*}

\begin{itemize}[itemsep=0 pt, topsep=0 pt]
   \setlength{\parsep}{0pt}
   \setlength{\itemsep}{0.5ex}
   \item w/GCN: replace Multi-view spatial attention(MVSA) module with GCN.
      \item w/o GTSA: remove the gated temporal self-attention .
   \item w/o LGSA: remove the local geographic self-attention .
   \item w/o GSSA: remove the global semantic self-attention .
   \item w/o PNSA: remove the pivotal nodes self-attention.
   \item w/o STCB: remove the spatio-temporal context broadcasting(STCB).
 \end{itemize}

The ablation experiment results are shown in Figure. \ref{figs/curve} and the observations from the results are as follows:
\begin{enumerate}
    \item Replacing MVSA with GCN results in a significant decline in performance. This finding indicates that GCN is not as effective as our MVSA in capturing spatial dependencies. The superior performance of MVSA highlights its effectiveness in modeling the multi-scale spatial dependencies within the data.

    \item Modeling temporal correlations is crucial for improving prediction accuracy. Removing GTSA leads to a significant increase in prediction errors. This indicates that the module effectively captures temporal correlations by simultaneously considering both local and global relationships of temporal features. Additionally, MAE is often influenced by outliers, yet our MAE shows considerable improvement, suggesting that the module accurately predicts outliers, further highlighting its superiority.

    \item Removing both LGSA and GSSA leads to the model's ineffective capture of local and global spatial dependencies, leading to decreased performance. 
This proves the critical role of capturing both local and global spatial dependencies in traffic prediction. 
 Not considering pivotal nodes results in poorer model performance compared to LVSTformer, because pivotal nodes possess more complex spatio-temporal dependencies than ordinary nodes. 
By focusing on pivotal nodes and assigning them higher weights, LVSTformer can more effectively model spatio-temporal dependencies of pivotal nodes, thereby improving prediction capability.

    \item This indicates that the STCB module effectively alleviates the problem of the model overly focusing on a few key features or neglecting other valuable information due to attention scores being too high or too low. By manually inserting uniform attention between two layers of the model, it ensures reasonable allocation of attention scores. This enhances prediction accuracy and strengthens the model's robustness and generalization ability.
\end{enumerate}


 

\subsection{Long Term Prediction Performance(RQ3)}
We conduct experiments on the PeMS04 dataset to forecast traffic conditions for the next two hours, assessing the long-term prediction performance of LVSTformer. Table \ref{longterm} presents comparison results between our model and models that perform well in long-term prediction. 
We observe that LVSTformer consistently outperforms the baseline models across all time steps, and the performance gap between LVSTformer and the baseline methods gradually widens as the prediction horizon extends.
The superior performance of LVSTformer in long-term prediction further emphasizes its capability to capture and model complex spatio-temporal dependencies.

\subsection{Hyperparameter Sensitivity(RQ4)}
In this segment, we examine several hyperparameter sensitivity of LVSTformer ,including the number of transformer layers, learning rate, and hidden dimension.

\begin{figure*}
\centerline{\includegraphics[width=0.8\textwidth]{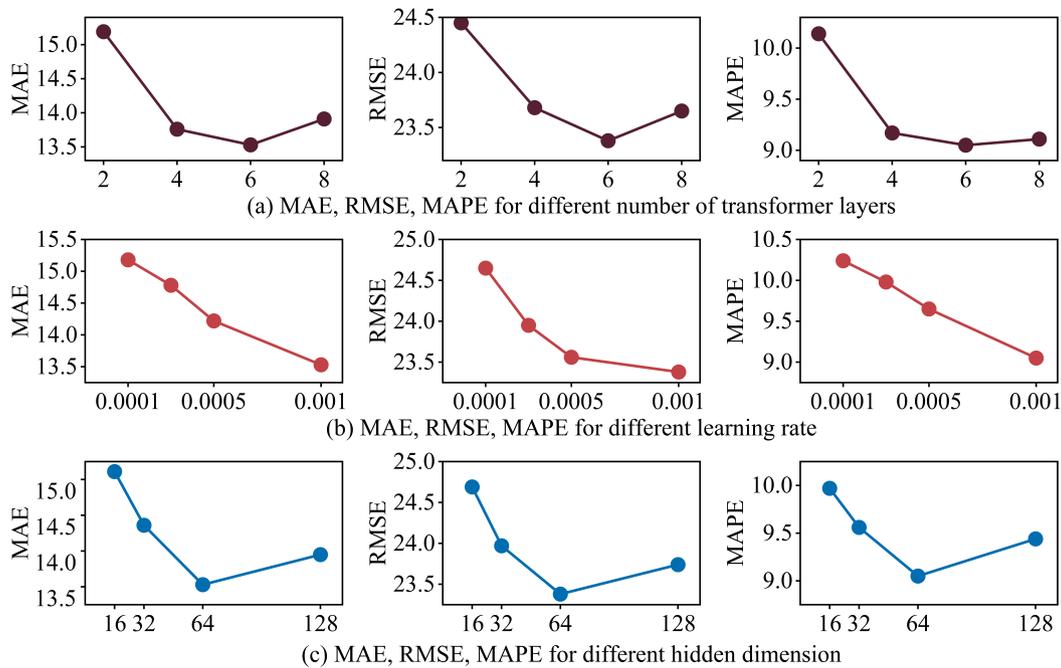}}
\caption{The parameter sensitivity experiments of LVSTformer with respect to three metrics (MAE, RMSE and MAPE) on PeMS08.}
\label{figs/parameter}
\end{figure*}

\textbf{The impact of transformer layers $L$}. We analyze the influence of
transformer layers by varying it within the range of  $\{2, 4, 6, 8\}$ while keeping
other parameters at their default settings. As shown in Figure \ref{figs/parameter}(a), optimal performance is attained with $L$ = 6. Further increasing the number of layers ($L$=8) resulted in a decline in performance, which suggests that six layers in our transformer architecture is sufficient to model spatio-temporal dependencies.
\textbf{The learning rate $\gamma$.}
We vary the learning rate $\gamma$ within the ranges of $\{10^{-5}, 3^{-4},  5^{-4}, 10^{-4}\}$. From Figure \ref{figs/parameter}(b), we observe a significant improvement in prediction accuracy as the learning rate increased from $10^{-5}$ to $10^{-4}$. Therefore, the model performance is optimal when the initial learning rate is set to $10^{-4}$. 

\textbf{The impact of hidden dimension $l$.}
As shown in Figure \ref{figs/parameter}(c), we vary hidden dimension  within the ranges of $\{16, 32, 64, 128\}$, the best performance of our
LVSTformer is achieved when $l=$64. However, the further increasing of $l$ leads to
the slightly performance degradation.
\begin{figure}
\centerline{\includegraphics[width=0.5\textwidth]{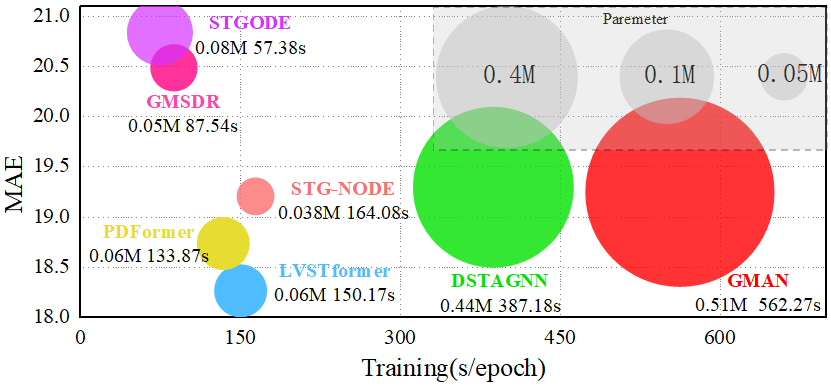}}
\caption{Time cost of different approaches.}
\label{figs/cost}
\end{figure}

\subsection{Computation Cost(RQ5)}
In this section, we evaluate number of parameters and training time for LVSTformer and six SOTA baseline methods using the same NVIDIA RTX 3090 GPU on the PEMS04 dataset. The comparison results are illustrated in Figure \ref{figs/cost}, it is evident that LVSTformer offers a more cost-effective choice in terms of training time and parameter efficiency compared to other methods, while still achieving optimal performance. 
Specifically, LVSTformer reduces the number of parameters by 7-8 times compared to DSTAGNN and decreases training time by 61.27\%. Additionally, LVSTformer reduces the number of parameters by 8-9 times compared to GMAN, resulting in a reduction in training time by 73.29\%. Compared to both of them, our model requires fewer computational resources and memory while maintaining optimal performance.
Although PDFormer has similar parameter count and training time to LVSTformer, it utilizes the k-shape clustering algorithm to model network delay, which requires significant computational resources. Additionally, while STGODE and GMSDR have fewer parameters and training time, LVSTformer significantly outperforms them in terms of prediction performance. These findings indicate that LVSTformer achieves a balance between efficiency and accuracy.

\subsection{Visualization of Multi-view Spatial Attention}
\begin{figure*}
\centerline{\includegraphics[width=1.0\textwidth]{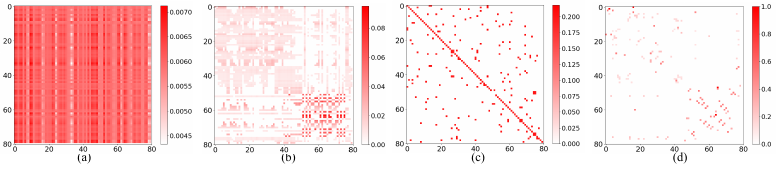}}
\caption{
For clarity, we selected the first 80 nodes of PeMS08 for visualization. Among them, (a) represents ordinary spatial self-attention, (b) represents local geographic self-attention, (c) represents global semantic self-attention, and (d) represents the pivotal node self-attention before the allocation of computational weights.}
\vspace{-1em}
\label{figs/attention}
\end{figure*}

In this section, we aim to demonstrate the effectiveness of multi-view spatial attention and enhance its interpretability. We compare and analyze the attention maps of LVSTformer from three different perspectives.
When spatial attention is not fused with spatial enhancement views, as shown in Figure \ref{figs/attention} (a), the attention is widely dispersed, with almost all nodes sharing the model's attention. This scattered attention indicates that the model's focus is not concentrated on specific regions, which can weaken its performance.
However, when the spatial attention is fused with certain spatial enhancement views, as depicted in Figure \ref{figs/attention} (b), (c) and (d), the scope of spatial attention becomes more constrained and focused on specific regions. For example, Figure \ref{figs/attention} (b) focuses on local geographic spatial feature, Figure \ref{figs/attention} (c) focuses on global semantic spatial feature and Figure \ref{figs/attention} (d) focuses on certain pivotal nodes.
This indicates the effectiveness of multi-view spatial attention in capturing local, global, and pivotal node spatial correlations. By incorporating different spatial enhancement views, the model can better identify and attend to relevant spatial information, leading to improved performance.

\section{Conclusion}
In this study, we propose a multi-level multi-view spatio-temporal Transformer model, namelby LVSTformer, to caputure rich spatial and temporal features for traffic prediction. Specifically, we first develop a spatio-temporal data embedding layer to aggregate raw traffic data, temporal periodic features, and spatial features, better integrating the spatio-temporal characteristics of traffic data. We then design a novel multi-level spatio-temporal transformer model composed of MVSA and GTSA. MVSA combines spatial enhancement views with spatial attention, enhancing the learning ability of spatial features. GTSA focuses on extracting local and global temporal features, enhancing the model's ability to capture short and long-term temporal dependencies.
Finally, we introduce spatio-temporal context broadcasting, manually inserting uniform attention between two layers of the model. This technique ensures a well-distributed allocation of attention scores, mitigating overfitting and information loss. 
We conduct extensive experiments on six datasets, further demonstrating the excellent performance of our model in traffic prediction. In future work, we plan to apply LVSTformer to other spatio-temporal prediction tasks beyond traffic prediction.

\section{Acknowledgements}
\par This work was supported by the China Postdoctoral Science Foundation under Grant No.2022M711088.

\bibliographystyle{apalike}
\bibliography{ref}

\end{document}